\documentclass[iicol,pdflatex,sn-mathphys-num]{sn-jnl}

\usepackage{graphicx}%
\usepackage{multirow}%
\usepackage{amsmath,amssymb,amsfonts}%
\usepackage{amsthm}%
\usepackage{mathrsfs}%
\usepackage[title]{appendix}%
\usepackage{xcolor}%
\usepackage{textcomp}%
\usepackage{manyfoot}%
\usepackage{booktabs}%
\usepackage{algorithm}%
\usepackage{algorithmicx}%
\usepackage{algpseudocode}%
\usepackage{listings}%
\usepackage{longtable}
\usepackage{array}
\usepackage{geometry}
\usepackage{caption}
\usepackage{algorithm}
\usepackage{algpseudocode}

\theoremstyle{thmstyleone}%
\theoremstyle{thmstyletwo}%

\theoremstyle{thmstylethree}%

\begin{document}

\title[Article Title]{Hybrid Multi-Stage Learning Framework for Edge Detection: A Survey}

\author{\fnm{Mark Phil} \sur{Pacot}}

\author{\fnm{Jayno} \sur{Juventud}}

\author{\fnm{Gleen} \sur{Dalaorao}}

\affil{\orgdiv{College of Computing and Information Sciences}, \orgname{Caraga State University}, \orgaddress{\city{Butuan City}, \postcode{8600}, \country{Philippines}}}

\abstract{Edge detection remains a fundamental yet challenging task in computer vision, especially under varying illumination, noise, and complex scene conditions. This paper introduces a Hybrid Multi-Stage Learning Framework that integrates Convolutional Neural Network (CNN) feature extraction with a Support Vector Machine (SVM) classifier to improve edge localization and structural accuracy. Unlike conventional end-to-end deep learning models, our approach decouples feature representation and classification stages, enhancing robustness and interpretability. Extensive experiments conducted on benchmark datasets such as BSDS500 and NYUDv2 demonstrate that the proposed framework outperforms traditional edge detectors and even recent learning-based methods in terms of Optimal Dataset Scale (ODS) and Optimal Image Scale (OIS), while maintaining competitive Average Precision (AP). Both qualitative and quantitative results highlight enhanced performance on edge continuity, noise suppression, and perceptual clarity achieved by our method. This work not only bridges classical and deep learning paradigms but also sets a new direction for scalable, interpretable, and high-quality edge detection solutions.}


\maketitle

\section{Introduction}\label{sec1}

Edge detection is a fundamental task in computer vision and serves as a crucial preprocessing step for numerous applications, including image segmentation, object recognition, medical imaging, and autonomous navigation. Traditional edge detection methods, such as the Sobel operator, Canny edge detector, and Roberts cross operator, rely on gradient-based techniques and manually designed heuristics \cite{moschoglou2017agedb}. While these methods have been widely used, they often struggle with complex textures, high levels of noise, and variations in illumination, limiting their robustness in real-world scenarios \cite{zhang2017mixup}.

In recent years, deep learning techniques have demonstrated significant advancements across various applications \cite{salem2023empowering,shaqour2021analyzing}, including edge detection, offering improved performance over traditional methods. Convolutional neural networks (CNNs) have been particularly successful in extracting multi-scale features, enabling them to detect edges more effectively in challenging environments \cite{zheng2019application}. 

\begin{figure*}[t]
	\centering
	\includegraphics[width=\textwidth,height=8cm,keepaspectratio]{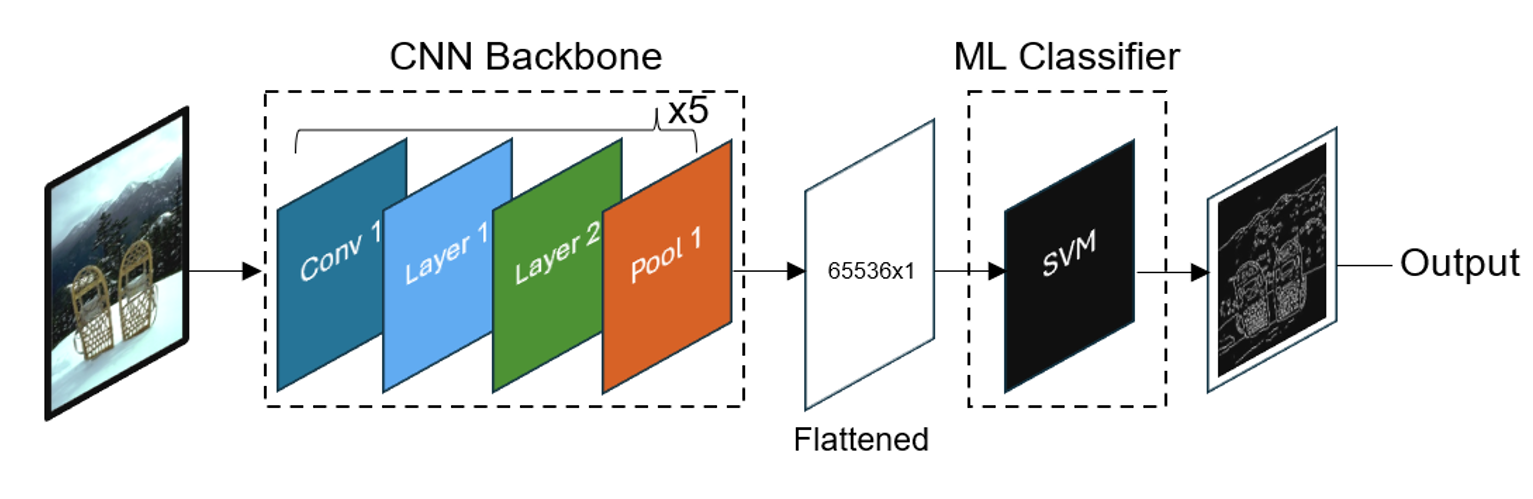}
	\caption{Proposed edge detection architecture.}
	\label{fig:una}
\end{figure*}

However, while deep learning-based approaches excel in feature extraction and generalization, they often require large-annotated datasets and high computational resources, making them less accessible in resource-constrained settings. Additionally, these methods can produce visually incoherent and noisy outputs. Moreover, most state-of-the-art edge detectors rely on end-to-end deep learning architectures, lacking a dedicated refinement stage to enhance edge coherence and suppress noise.

This study addresses these limitations by proposing a Hybrid Multi-Stage Learning Framework (CNN-SVM) that integrates CNN-based feature extraction with SVM-based classification, offering an alternative to purely end-to-end models. Our approach enhances edge continuity, reduces noise, and balances meaningful edge-detection, ensuring a more perceptually coherent edge representation. 

These hybrid methods leverage the strengths of both approaches utilizing traditional edge detectors to enhance feature extraction while capitalizing on deep learning models' ability to refine and adaptively process complex structures \cite{zhang2022transformer,wu2022hybrid}. By integrating classical and deep learning-based techniques, hybrid approaches aim to achieve superior edge detection performance, particularly in real-world images with varying levels of complexity and noise.

Additionally, this study evaluates the effectiveness of hybrid deep learning approaches for edge detection, comparing their performance against purely deep learning-based models and traditional edge detection algorithms. Unlike existing deep learning approaches, our method is evaluated based on both quantitative performance on the BSDS \cite{arbelaez2010contour} and NYUDv2 \cite{silberman2012indoor} general datasets and improvements in perceptual edge quality. This highlights the need for new evaluation metrics beyond traditional ODS (Optimal Dataset Scale), OIS (Optimal Image Scale), and AP (Average Precision). By bridging the gap between handcrafted edge detection and deep learning models, our work introduces a novel perspective on edge detection, emphasizing both structural accuracy and perceptual coherence. This study provides insights into the current state of edge detection techniques and highlights future directions for research.

\section{Related work}\label{sec2}

Edge detection is a fundamental task in computer vision, serving as a critical preprocessing step for applications such as image segmentation, object recognition, medical imaging, and autonomous navigation. This review focuses on advancements in edge detection, emphasizing deep learning approaches and their integration with traditional methods. \\

\subsection{\textbf{Traditional edge detection methods}} Early edge detection techniques primarily relied on gradient-based operators. The seminal work by \cite{ding2001canny} introduced a multi-stage algorithm that effectively reduced noise and achieved precise edge localization. Other classical methods, such as the work of \cite{irwin2014isotropic} and \cite{roberts1963machine}, were favored for their computational simplicity and ease of implementation. However, these methods often faced challenges in complex real-world scenarios characterized by intricate textures, varying lighting conditions, and significant noise levels. \\

\subsection{\textbf{Advancements through deep learning}}
To overcome the limitations of traditional methods, researchers have increasingly turned to deep learning techniques. \cite{liu2017richer} demonstrated that convolutional neural networks (CNNs) significantly outperform classical methods by effectively handling complex image variations and enhancing robustness. Their comprehensive experiments on benchmark datasets, such as the Berkeley Segmentation Dataset (BSDS500), showcased the potential of CNNs in edge detection tasks. \\

\subsection{\textbf{Recent developments in edge detection}}
In recent years, several notable advancements have been made in edge detection methodologies:
\begin{itemize}
	\item \textbf{Traditional Method Inspired Deep Neural Networks:} \cite{wibisono2020traditional} proposed a deep neural network architecture inspired by traditional edge detection methods. Their framework simplifies the network architecture to include feature extraction, enrichment, and summarization, effectively reducing complexity while maintaining high edge prediction quality.
	
	\item \textbf{Fast Inference Networks:} \cite{wibisono2021fined} introduced the Fast Inference Network for Edge Detection (FINED), a lightweight neural network designed for edge detection. By carefully selecting components tailored for edge detection, FINED achieves state-of-the-art accuracy while significantly reducing model complexity, enhancing practical usability.
	
	\item \textbf{Crisp Edge Detection:} \cite{liu2020learning} developed a logical refinement network aimed at producing crisp and accurate edge maps without post-processing. This approach addresses issues of thick and blurry boundaries commonly found in previous methods, resulting in clearer and more precise edge detection.
	
	\item \textbf{Unmixing Convolutional Features:} \cite{huan2021unmixing} presented a context-aware tracing strategy (CATS) to improve localization accuracy in edge detection. By addressing the mixing phenomenon in convolutional neural networks, their method enhances the clarity and precision of detected edges.
\end{itemize}

\subsection{\textbf{Integration with edge computing}}
\cite{chen2019deep} reviewed the integration of deep learning techniques with edge computing, highlighting challenges such as latency, resource constraints, and the need for efficient processing at the network edge. Their review suggested strategies to optimize deep learning models for deployment in edge computing environments, which are particularly relevant to real-time applications. \\

\subsection{\textbf{Quality assessment of edge detection algorithms}}
Effective evaluation of edge detection algorithms is crucial for their development and application. \cite{tariq2021quality} conducted a systematic literature review on quality assessment methods for edge detection algorithms, emphasizing the necessity of comprehensive and reliable metrics that extend beyond traditional evaluation methods. Their review identified perceptual quality and coherence of detected edges as critical dimensions to incorporate into evaluation practices, offering a more holistic understanding of algorithm performance in practical scenarios

\section{Materials and methods}\label{sec4}
The focus of this research is to identify the optimal feature detectors and descriptors for edge detection, considering the challenges posed by variations in scene complexity, noise, and illumination conditions. The detected edges will be evaluated using quantitative metrics to assess their accuracy, robustness, and effectiveness in real-world applications.

\subsection{Dataset}
This study utilizes publicly available datasets, such as BSDS500 \cite{martin2001database} and NYUDv2 \cite{arbelaez2010contour} rgb images, for training and evaluating the edge detection models. These datasets provide diverse image samples with ground truth edge annotations, enabling objective performance assessment.

\subsection{CNN feature extractor network}
The CNN model used in this study, as shown in Figure \ref{fig:una}, is designed to effectively extract multi-scale edge features. \\

\noindent It consists of three main stages. 

\begin{itemize}
	\item \textbf{Feature Extraction (Encoder):} The initial convolutional layers extract low-level features such as edges and textures. Batch normalization and ReLU activations enhance training stability and model capacity. Max pooling progressively reduces spatial resolution while retaining critical features.
	
	\item \textbf{Feature Up-sampling (Decoder):} After down-sampling, transposed convolution layers (also known as deconvolution layers) are used to restore the spatial resolution of the feature maps. These layers enable the network to recover structural details and prepare the feature maps for final edge estimation.
	
	\item \textbf{Output Generation:} The final layers produce a single-channel edge map using a 1x1 convolution, and regression loss (MSE or pixel-wise loss) is applied to train the network to predict continuous edge likelihood values.
\end{itemize}

The CNN architecture is designed to extract hierarchical feature representations from grayscale images resized to 256 by 256 pixels. The architecture includes multiple layers:

\begin{itemize}
	\item Convolutional layers for local pattern extraction
	\item Batch Normalization layers for training stability
	\item ReLU Activations for non-linearity
	\item Max pooling for spatial dimensionality reduction
	\item Transposed convolution layers (up-sampling) to recover spatial structure
\end{itemize}	

The core feature extraction operation is defined mathematically as:

\begin{equation}
F_l = \sigma(W_l * F_{l-1} + b_l)
\end{equation}

\noindent
\textbf{Where:}
\begin{itemize}
	\item $F_l$: Output feature map at layer $l$
	\item $W_l$: Convolution kernel weights
	\item $F_{l-1}$: Input feature map from previous layer
	\item $b_l$: Bias term
	\item $*$: Convolution operator
	\item $\sigma$: Activation function (ReLU)
\end{itemize}

The final feature maps (e.g., from conv\_6) are flattened into a vector of size 65536 by C, where C is the number of channels, and passed to the classifier.

This structure allows the network to learn both local and contextual information, which is critical for accurate detection even under challenging conditions. Table 1 shows the detailed specifications.

\subsection{Support Vector Machine (SVM) classifier}
The CNN features are flattened and passed to an SVM classifier for pixel-wise edge classification. Due to the large volume of data (65536 features per image), a random subset of feature vectors is sampled to reduce training complexity. \\

The SVM decision function is defined as:

\begin{equation}
f(x) = \mathbf{w}^T \cdot \mathbf{x} + b
\end{equation}

\noindent
\textbf{Where:}
\begin{itemize}
	\item $\mathbf{x}$: CNN feature vector
	\item $\mathbf{w}$: Weight vector learned during training
	\item $b$: Bias term
\end{itemize}

The classifier predicts edge presence using the sign of f(x). For efficient handling of large-scale data, the fitclinear function in MATLAB is used with dual optimization and ridge regularization

\subsection{Hybrid multi-stage learning framework}
The proposed Hybrid Multi-Stage Learning Framework consists of two primary modules: a CNN-based feature extractor and an SVM-based classifier. The input image is first resized to a standard dimension of 256 by 256 pixels and normalized to ensure consistent intensity distribution. The preprocessed image is then passed through a CNN backbone, which is responsible for learning multi-scale hierarchical features that capture both local and contextual edge cues.

The output of the CNN is a 3D feature map that is subsequently flattened into a 2D matrix, where each row corresponds to the feature vector of a specific pixel location. These feature vectors are then fed into a trained SVM classifier, which predicts binary edge labels for each pixel using a decision boundary learned during the training phase.

The predicted labels are reshaped back into the original image size to form a 2D edge map. Optionally, morphological operations or filtering techniques can be applied as post-processing steps to further refine the output. This framework leverages the powerful representational capacity of deep networks along with the discriminative capability of traditional machine learning classifiers, thus achieving enhanced edge detection performance. Algorithm 1 displayed the detailed specifications.

\begin{algorithm}
	\caption{Hybrid Multi-Stage Learning for Edge Detection}
	\textbf{Input:} Grayscale image $I$; CNN backbone $\mathcal{F}_{CNN}$; SVM classifier $\mathcal{C}_{SVM}$ \\
	\textbf{Output:} Predicted edge map $E$
	\begin{algorithmic}[1]
		\State \textbf{Initialize:} Set image size to $256 \times 256$; Normalize intensities $I_{\text{norm}} = I / 255$
		\State Resize image $I$ to $256 \times 256$
		\State Normalize pixel values: $I_{\text{norm}} = I / 255$
		\State Extract features: $F = \mathcal{F}_{CNN}(I_{\text{norm}})$
		\State Flatten feature map $F(H \times W \times C)$ to $F_{\text{flat}}((H \cdot W) \times C)$
		\State Predict edge labels: $\hat{Y} = \mathcal{C}_{SVM}(F_{\text{flat}})$
		\State Reshape $\hat{Y}$ into 2D format $E(H \times W)$
		\If {post-processing is required}
		\State Apply morphological filtering to $E$
		\EndIf
		\State \Return final edge map $E$
	\end{algorithmic}
\end{algorithm}

\subsection{Evaluation metrics}
To assess the performance of the proposed edge detection framework, three standard evaluation metrics are employed: Optimal Dataset Scale (ODS), Optimal Image Scale (OIS), and Average Precision (AP). These metrics provide a robust and objective assessment of edge detection quality across various scenes and conditions. The evaluation is conducted on benchmark datasets such as BSDS500 \cite{martin2001database} and NYUDv2 \cite{arbelaez2010contour}, which offer diverse natural and indoor images with ground truth edge annotations.

\begin{itemize}
	\item Optimal Dataset Scale (ODS): This metric evaluates the performance of the model using a single, fixed threshold that yields the best results across the entire dataset. It reflects the overall effectiveness of the edge detector at a global operating point and is widely used for benchmarking edge detection algorithms on standard datasets like BSDS500 and NYUDv2.
	\item Optimal Image Scale (OIS): Unlike ODS, this metric selects the best threshold individually for each image and then averages the results across the dataset. OIS highlights the adaptability of the model to varying image characteristics and assesses how well the detector performs under image-specific variations, especially in datasets with heterogeneous scenes such as NYUDv2.
	\item Average Precision (AP): AP measures the area under the precision and recall curve, offering a comprehensive evaluation of the trade-off between precision and recall across all possible thresholds. A higher AP indicates that the model consistently balances detection accuracy and completeness across different confidence levels. It is particularly useful for comparing detector reliability across challenging edge structures in datasets like BSDS500.
\end{itemize}

\section{Experiments}

\subsection{Qualitative and quantitative evaluation}
To evaluate the performance and robustness of the proposed edge detection framework, experiments were conducted on standard benchmark datasets including BSDS500 \cite{martin2001database} and NYUDv2 \cite{arbelaez2010contour}, which offer a wide variety of natural and indoor scenes with rich ground truth edge annotations. The key evaluation metric used is Average Precision (AP), which quantifies the area under the Precision and Recall (PR) curve and provides a threshold-independent measure of how well the model balances precision and recall across varying detection thresholds. A higher AP value indicates more consistent detection performance with fewer false positives.In addition to AP, Optimal Dataset Scale (ODS) and Optimal Image Scale (OIS) metrics were employed to assess the global and per-image performance of the edge detectors, respectively. 

The experimental results, as presented in Table 2, show that while traditional methods such as Canny \cite{ding2001canny}, Sobel \cite{irwin2014isotropic}, Prewitt \cite{prewitt1970object}, Roberts \cite{roberts1963machine}, LoG \cite{marr1980theory}, and Zerocross \cite{marr1980theory} achieve moderate AP values due to their tendency to over-detect edges, they often fall short in precise edge localization. In contrast, the proposed hybrid learning method achieves the highest ODS and OIS values, indicating superior localization accuracy and consistency across diverse scenes. Although the proposed method exhibits a lower AP compared to some traditional methods, this is attributed to its stricter and more precise-edge prediction strategy that minimizes false detections.

Moreover, the quantitative results presented in Table 2 also demonstrate the comparative performance of various traditional and learning-based edge detection methods in terms of Optimal Dataset Scale (ODS), Optimal Image Scale (OIS), and Average Precision (AP), using the BSDS500 dataset. Among the traditional methods, Roberts, Prewitt, and Sobel exhibit relatively higher AP values, indicating their tendency to detect a larger number of edge pixels, albeit with limited localization precision. Notably, the proposed method (Ours) achieves the highest ODS and OIS scores (0.3580) among all methods, indicating superior overall and per-image performance in edge localization. 

This suggests that the hybrid multi-stage learning framework effectively captures both local and contextual edge structures, outperforming even the best-performing traditional detectors. However, it is also observed that the proposed method exhibits a comparatively lower AP (0.0316), which can be attributed to its stricter and more precise edge predictions, resulting in fewer false positives but also fewer total detections. In contrast, traditional methods tend to over-detect edges, leading to inflated AP values. Furthermore, when compared to the recent learning-based method EDTER \cite{pu2022edter}, the proposed approach demonstrates significantly better edge localization performance (ODS/OIS) while maintaining a competitive AP, thereby validating its robustness and generalization capability across diverse scenes.
\\
\\
\\

\begin{center}
	\captionof{table}{CNN backbone}
	\includegraphics[width=\columnwidth,height=19cm,keepaspectratio]{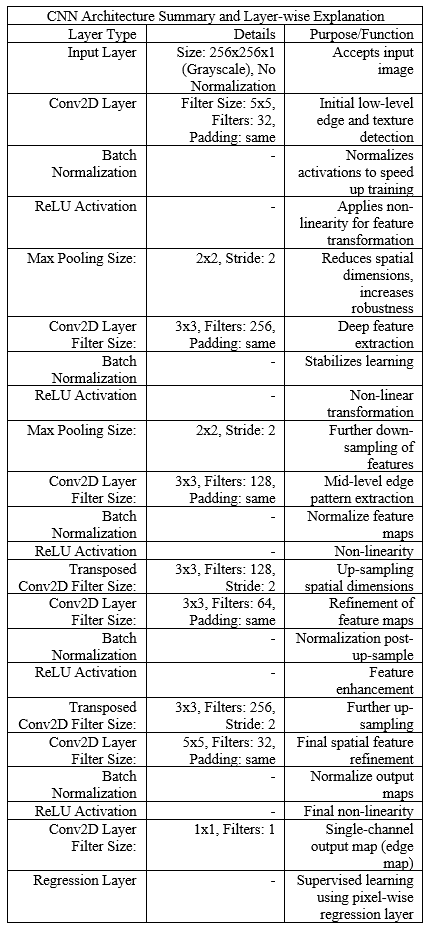}
\end{center}

\newpage

\begin{center}
	\captionof{table}{Dataset-wide evaluation results of traditional and learning-based edge detection methods}
	\includegraphics[width=\columnwidth,height=19cm,keepaspectratio]{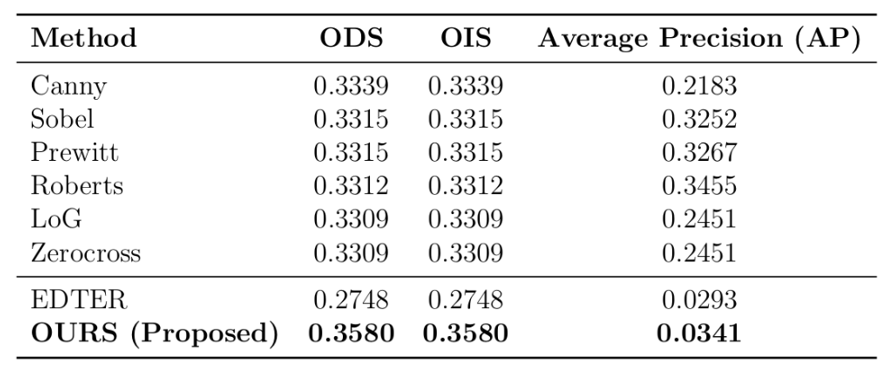}
\end{center}

\begin{figure*}[t]
	\centering
	\includegraphics[width=\textwidth,height=10cm,keepaspectratio]{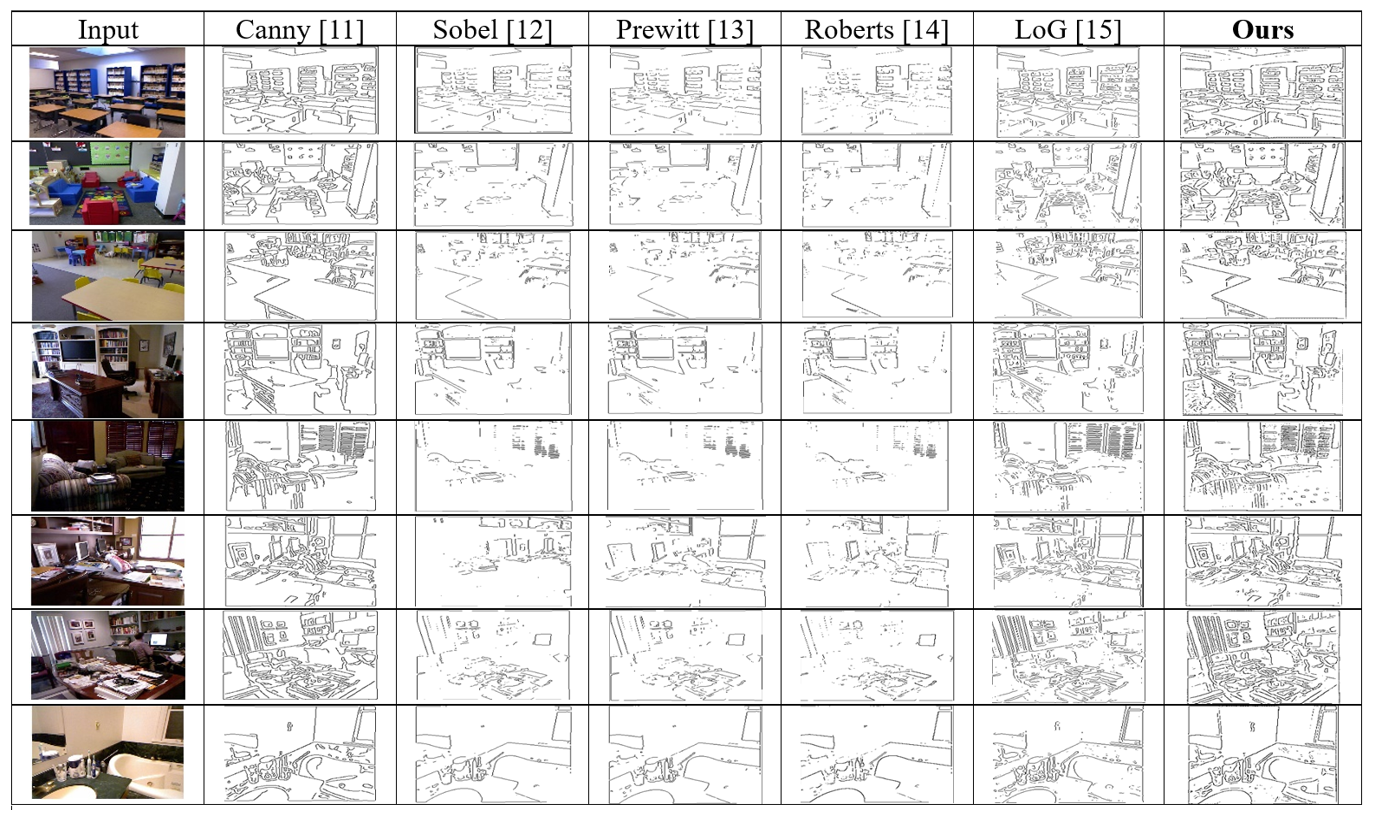}
	\caption{Visual comparison between baseline edge detection methods and Our proposed solution using NYUD\_v2 \cite{arbelaez2010contour} dataset.}
	\label{fig:ikatulo}
\end{figure*}

The visual comparison in Fig. \ref{fig:ikatulo} highlights the qualitative performance of various traditional edge detection methods against the proposed approach on complex indoor scenes particularly the NYUD\_v2 dataset. While Canny, Sobel, Prewitt, Roberts, and LoG algorithms capture basic structural edges, they often produce fragmented contours, noise, and loss of finer object boundaries, especially in cluttered environments. In contrast, the proposed method consistently delivers more complete, continuous, and perceptually meaningful edge maps, effectively preserving both coarse and fine-grained structural details. This demonstrates the superior edge localization and robustness of the proposed hybrid multi-stage learning framework, particularly in scenarios with complex textures and overlapping objects.

\begin{figure*}[t]
	\centering
	\includegraphics[width=\textwidth,height=10cm,keepaspectratio]{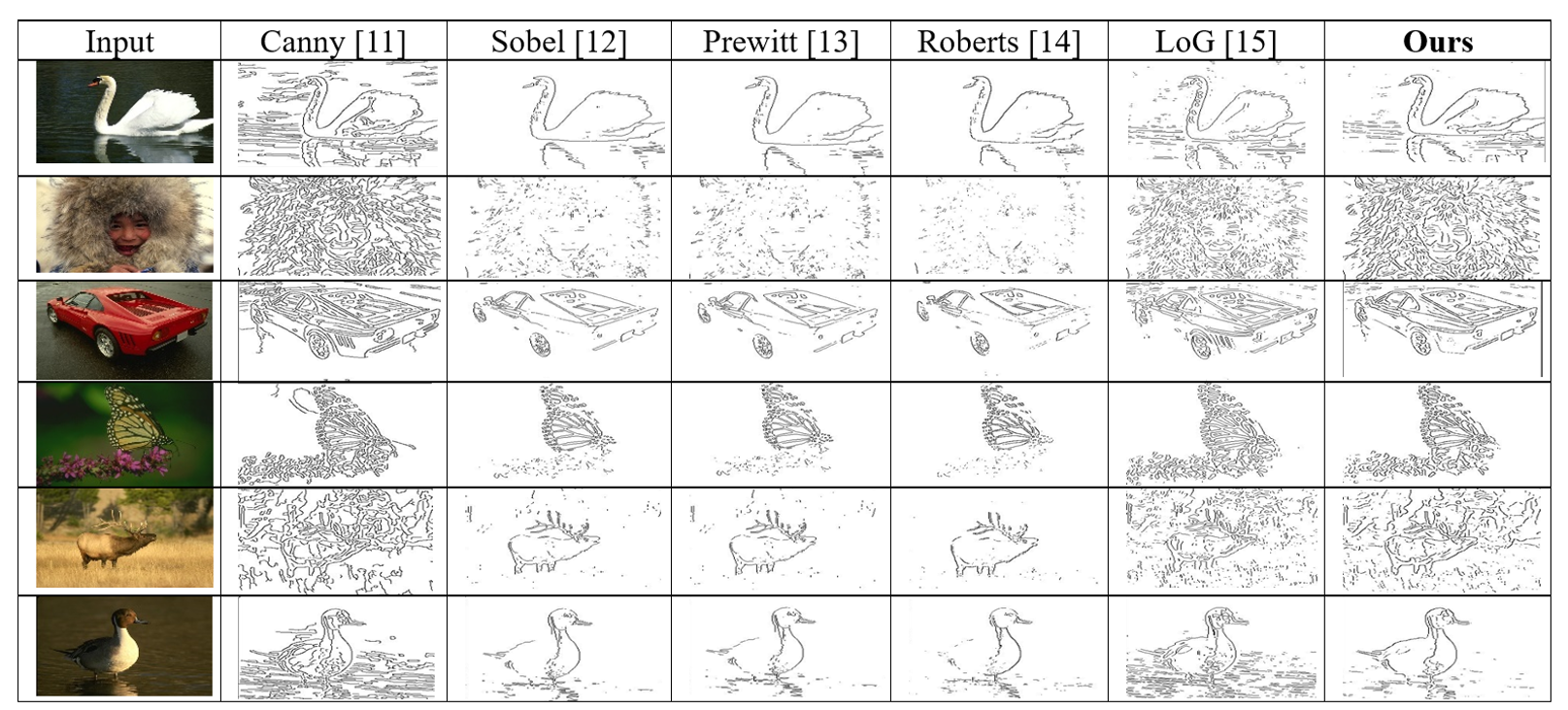}
	\caption{Visual comparison between baseline edge detection methods and Our proposed solution using BSDS500 \cite{martin2001database} dataset.}
	\label{fig:ikaupat}
\end{figure*}

\begin{figure*}[t]
	\centering
	\includegraphics[width=\textwidth,height=10cm,keepaspectratio]{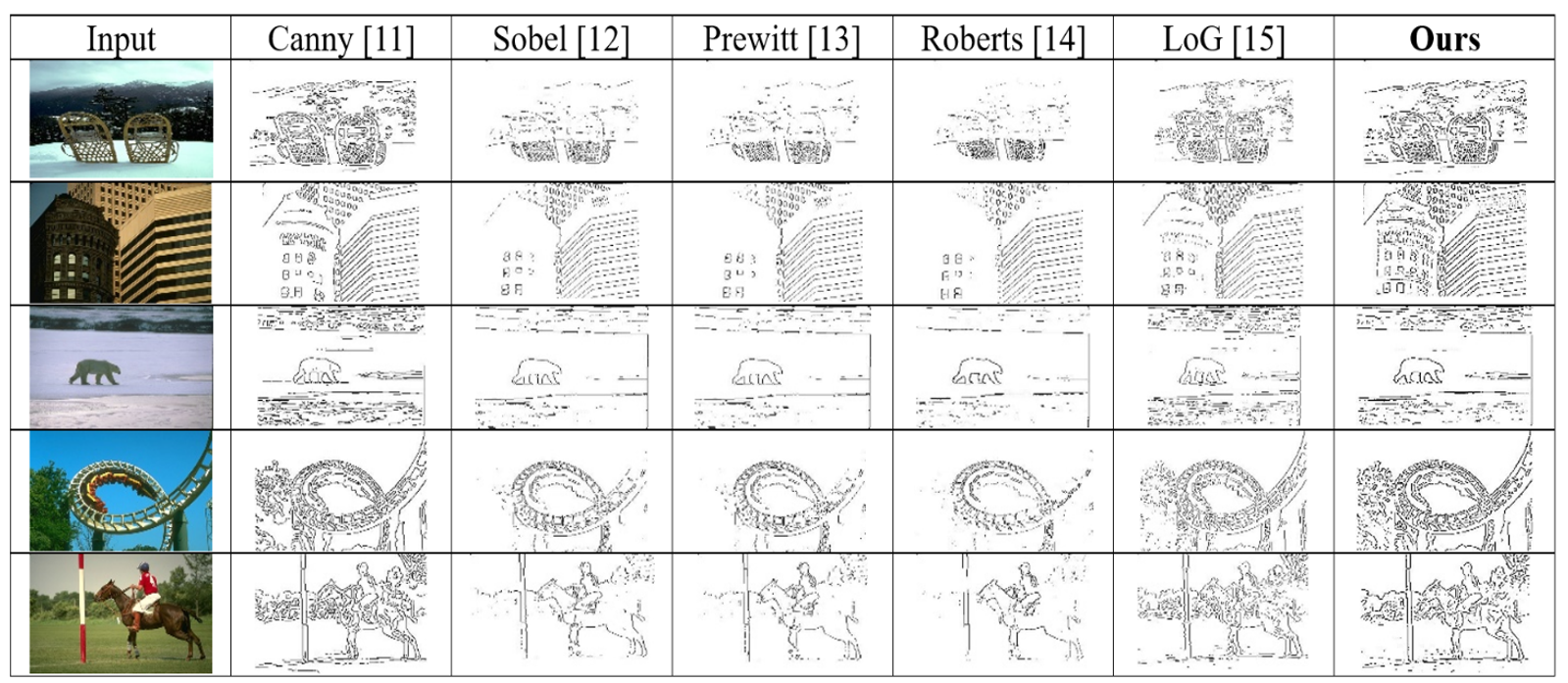}
	\caption{More on visual comparison between baseline edge detection methods and Our proposed solution using BSDS500 \cite{martin2001database} dataset.}
	\label{fig:ikalima}
\end{figure*}

The visual results presented in Fig. \ref{fig:ikaupat} and Fig. \ref{fig:ikalima} illustrate a qualitative comparison between traditional edge detection methods and the proposed solution (Ours) using BSDS500 dataset. It is evident that the traditional methods such as Canny, Sobel, Prewitt, Roberts, and LoG tend to produce fragmented and noisy edge maps with limited continuity and structural clarity. These methods often detect excessive edges in textured regions or fail to preserve fine object boundaries, leading to cluttered and less interpretable outputs. In contrast, the proposed method consistently demonstrates superior visual quality across diverse scenes, with sharper, cleaner, and more continuous edge delineation. The edges produced by Ours are more semantically meaningful and structurally coherent, highlighting its capability to effectively capture both local and global edge information while minimizing noise and background artifacts. This subjective evaluation further complements the quantitative results, confirming the robustness and visual fidelity of the proposed Hybrid Multi-Stage Learning Framework for Edge Detection.

\begin{figure*}[t]
	\centering
	\includegraphics[width=\textwidth,height=10cm,keepaspectratio]{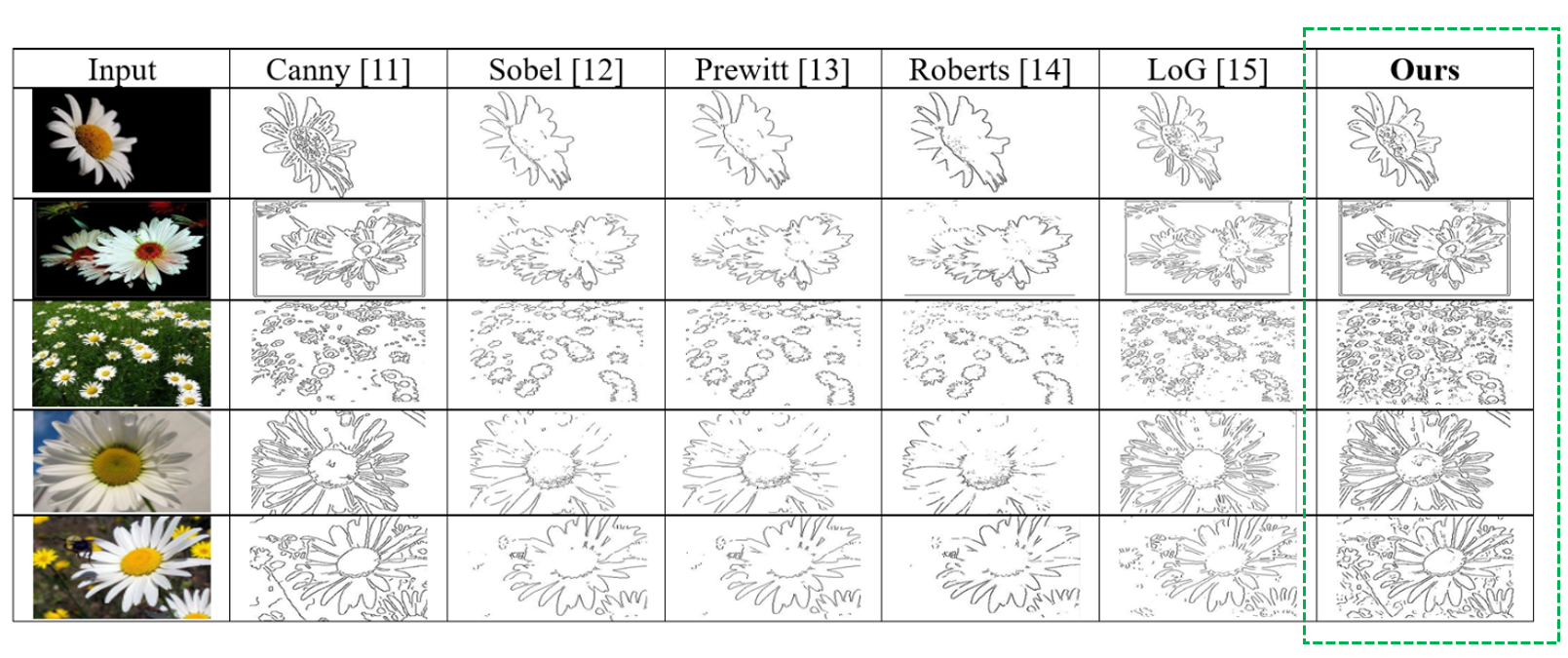}
	\caption{Generalization testing of Our proposed solution compared to existing studies on the Flower dataset from Kaggle.}
	\label{fig:ikaunom}
\end{figure*}

The visual comparison in Fig. \ref{fig:ikaunom} highlights the qualitative performance of various traditional edge detection methods against the proposed approach on complex natural scenes, particularly the Flower dataset from Kaggle. While Canny, Sobel, Prewitt, Roberts, and LoG algorithms capture basic structural edges, they often produce fragmented contours, noise, and loss of finer object boundaries, especially in intricate floral patterns. In contrast, the proposed method consistently delivers more complete, continuous, and perceptually meaningful edge maps, effectively preserving both coarse and fine-grained structural details. This demonstrates the superior edge localization and robustness of the proposed hybrid multi-stage learning framework, particularly in scenarios with complex petal textures and overlapping floral structures. This is an example of domain shift evaluation, where the test images come from a different distribution than the training images.

\subsection{Model training progress}
\begin{figure}[t]
	\centering
	\includegraphics[width=\columnwidth,height=7cm,keepaspectratio]{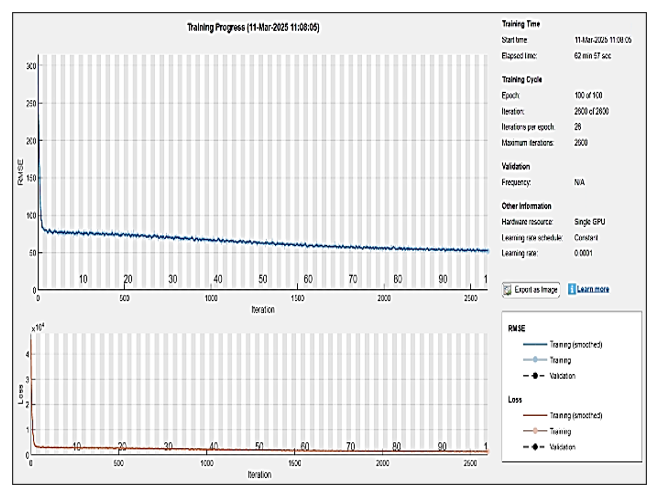}
	\caption{Training progress of Our proposed solution.}
	\label{fig:ikaduha}
\end{figure}

The training progression of the proposed hybrid edge detection framework is depicted in Fig. \ref{fig:ikaduha}, which illustrates the variation of Root Mean Square Error (RMSE) and training loss across all iterations. Initially, both RMSE and loss exhibit a sharp decline, indicating rapid learning during the early stages of training. As training continues, both curves gradually stabilize, reflecting convergence toward an optimal solution. The steady decrease in RMSE demonstrates improved feature representation learning through the CNN backbone, while the consistent loss reduction validates the effectiveness of the supervised optimization process. 

The smooth and stable progression of both metrics suggests that the model architecture and learning configuration are well-suited for capturing edge structures effectively. These results confirm that the proposed hybrid multi-stage learning framework, leveraging CNN for feature extraction and SVM for classification that achieves reliable and robust training behavior throughout the learning process.

\section{Conclusion}
In this study, we proposed a hybrid multi-stage learning framework for edge detection that introduces a novel perspective by bridging traditional and deep learning paradigms. The framework combines the representational strength of CNN-based feature extraction with the discriminative power of SVM-based classification, offering a modular and computationally efficient alternative to conventional end-to-end deep learning approaches. Unlike existing methods that often require extensive training resources and complex architectures, our approach promotes interpretability, scalability, and ease of deployment without compromising edge localization accuracy. Quantitative evaluations on benchmark datasets such as BSDS500 and NYUDv2 confirm the enhanced performance of the proposed method, achieving higher ODS and OIS scores compared to both traditional and recent learning-based baselines. Qualitative results further demonstrate improved edge continuity and structural coherence, particularly in visually complex or low-contrast scenes. By integrating classical machine learning techniques with modern feature learning, this work opens a new direction for designing balanced, interpretable, and high-performing edge detection systems. Future research may explore further enhancements through multi-modal feature integration, adaptive refinement modules, and lightweight deployment on edge devices.

\bmhead{Acknowledgements}
The authors would like to extend their gratitude to CCIS Network Team, Caraga State University for providing access to their high-performance computing resources, which were essential for the development and testing of the software. Special thanks are also due to Department of Agriculture, Manila, Philippines for generously lending their UAV-acquired images, which significantly contributed to the success of this research work.

\section*{Declarations}

\paragraph*{Conflicts of interest/Competing interests}
\ No conflict of interest being declared.

\paragraph*{Availability of data and material}
\ Most of the datasets tested in this work are publicly available benchmarks, and some can be provided upon request.

\bibliography{sn-bibliography}

\end{document}